\documentclass[11pt, twoside]{article}

\usepackage[usenames,dvipsnames,svgnames,table]{xcolor}
\usepackage[T1]{fontenc}
\usepackage{arxiv}

\usepackage{graphicx}
\usepackage{amsmath}
\usepackage{amssymb}
\usepackage[ruled,vlined]{algorithm2e}
\usepackage{lineno}
\usepackage{hyperref}
\hypersetup{colorlinks,allcolors=black}
\usepackage{ dsfont }
\usepackage{wrapfig}
\usepackage{makecell}
\usepackage{subcaption}
\usepackage{changepage,threeparttable}

\usepackage{booktabs, multirow} 
\usepackage{soul}
\usepackage[inline]{enumitem}

\usepackage{tablefootnote}
\let\tmp\oddsidemargin
\let\oddsidemargin\evensidemargin
\let\evensidemargin\tmp
\reversemarginpar

\fancypagestyle{plain}{
	
	\fancyfoot[C]{\twopagenumbers}
}

\newcounter{pageaux}

\newcommand{\twopagenumbers}{%
	\stepcounter{pageaux}%
	\thepage
}
\makeatletter

\newcommand\blfootnote[1]{%
  \begingroup
  \renewcommand\thefootnote{}\footnote{#1}%
  \addtocounter{footnote}{-1}%
  \endgroup
}

\definecolor{scorecolor}{HTML}{8DD3C7}

\newcommand{\mytitle}{Can we infer the presence of Differential Privacy in Deep Learning models' weights? Towards more secure Deep Learning}
\newcommand{\shorttitle}{}
\newcommand{\mychapter}{}

\usepackage{array}
\newcolumntype{L}[1]{>{\raggedright\let\newline\\\arraybackslash\hspace{0pt}}m{#1}}
\newcolumntype{C}[1]{>{\centering\let\newline\\\arraybackslash\hspace{0pt}}m{#1}}
\newcolumntype{R}[1]{>{\raggedleft\let\newline\\\arraybackslash\hspace{0pt}}m{#1}}

\newcommand{\dani}[1]{\textcolor{black}{#1}}
\newcommand{\danii}[1]{\textcolor{black}{#1}}
\newcommand{\daniiiii}[1]{\textcolor{black}{#1}}

\usepackage[numbers]{natbib} 
\usepackage[inline]{enumitem} 
\usepackage{graphicx}
\usepackage{tikz}
\usepackage{makecell}
\usepackage{dsfont}
\usepackage{multirow}
\usepackage{verbatim}
\usetikzlibrary{mindmap,shadows}
\usepackage{amssymb}
\usepackage{tabularx}

\title{\mytitle}
\date{}

\author{
        Daniel Jiménez López $^{*\text{,a}}$ \\
        \And
	Nuria Rodríguez-Barroso $^{\text{a}}$ \\
	\And
        M. Victoria Luzón $^{\text{b}}$ \\
	\And
	Francisco Herrera $^{\text{a,c}}$ \\
}

\begin{document}

\maketitle

\vspace{-1cm}

\begin{centering}
$^\textbf{a}$ \textit{Department of Computer Science and Artificial Intelligence, Andalusian Research Institute in Data Science and Computational Intelligence (DaSCI), University of Granada, Spain}\\
$^\textbf{b}$ \textit{Department of Software Engineering, Andalusian Research Institute in Data Science and Computational Intelligence (DaSCI), University of Granada, Spain} 
\end{centering}

\blfootnote{* Corresponding Author}
\blfootnote{Email addresses: \textbf{\texttt{dajilo@ugr.es}} (Daniel Jiménez López), \textbf{\texttt{rbnuria@ugr.es}} (Nuria Rodríguez-Barroso),\textbf{\texttt{luzon@ugr.es}} (M. Victoria Luzón), \textbf{\texttt{herrera@decsai.ugr.es}} (Francisco Herrera)}

\vspace{0.5cm}

\begin{abstract}

Differential Privacy (DP) is a key property to protect data and models from integrity attacks. In the Deep Learning (DL) field, it is commonly implemented through the Differentially Private Stochastic Gradient Descent (DP-SGD). However, when a model is shared or released, there is no way to check whether it is differentially private, that is, it required to trust the model provider. This situation poses a problem when data privacy is mandatory, specially with current data regulations, as the presence of DP can not be certificated consistently by any third party. Thus, we face the challenge of determining whether a DL model has been trained with DP, according to the title question: \emph{Can we infer the presence of Differential Privacy in Deep Learning models’ weights?} Since the DP-SGD significantly changes the training process of a DL model, we hypothesize that DP leaves an imprint in the weights of a DL model, which can be used to predict whether a model has been trained with DP regardless of its architecture and the training dataset. In this paper, we propose to employ the imprint in model weights of using DP to infer the presence of DP training in a DL model. To substantiate our hypothesis, we developed an experimental methodology based on two datasets of weights of DL models, each with models with and without DP training and a meta-classifier to infer whether DP was used in the training process of a DL model, by accessing its weights. We accomplish both, the removal of the requirement of a trusted model provider and a strong foundation for this interesting line of research. Thus, our contribution is an additional layer of security on top of the strict private requirements of DP training in DL models, towards to DL models.

\end{abstract}


\clearpage

\section{Introduction}

The quick development and integration of Artificial Intelligence (AI) systems, as well as, the increased data collection enabled by Internet of Things devices and fast mobile networks such as 5G has started to significantly transform society. While offering great opportunities, AI systems also give rise to certain risks that must be handled appropriately \cite{DOYA2022542}. Particularly, privacy and transparency are key to protecting from the misuse of AI systems and deepening our understanding of them. In fact, they are two of the seven key elements required for Trustworthy AI \cite{gdpr2}.

As important as it is implementing measures to ensure privacy and transparency, it is certificating that such measures have been implemented by any third party. Accountability, a key property of Responsible AI systems \cite{BARREDOARRIETA202082} and more specifically, auditability contributes to trustworthiness of the technology. Responsible AI systems should be capable of being independently audited in applications affecting fundamental rights, including safety-critical applications.

With aims to balance the need to extract useful information from data while ensuring the privacy of individuals whose data is being analyzed, Differential Privacy (DP) is born \cite{dwork2014}. It creates a framework for designing privacy preserving mechanisms to access data and statistics. It is a useful tool in multiple fields of AI, specially Deep Learning (DL).

DL models are, by no means, secure and private by default, that is, they are susceptible to a wide range of privacy attacks \cite{overview_privacidad_centralizado}. DP has a well established extension to the DL field through the Differentially Private Stochastic Gradient Descent (DP-SGD) \cite{Abadi2016}. Thus, DL models can be made more private using the DP-SGD, enabling robust privacy protection for individuals. Still, it degrades the classification performance of the model, that is, it poses a trade-off between utility and privacy, which get worse in imbalanced scenarios \citep{bagdasaryan2019differential}.



To our knowledge, there is no way of checking if a model is DP, once the training phase ends. Consequentially, the model provider has to be trusted. This situation poses a problem in contexts where data privacy is a strict requirement, such as Machine Learning As a Service \citep{ribeiro2015mlaas} infrastructures where a DP model can be generated, but the presence of DP training can not be checked easily. Particularly, there is no way for a third party to certificate the enforcement of data privacy through DP of released DL models. Theoretically, Membership Inference Attacks \citep{carlini2021membership} can be used to estimate DP guarantees, but they fail at ensuring that a model has not been trained with DP, since they can not estimate properly the absence of DP \citep{measureDP}.


Considering the significant changes that the DP-SGD introduces in the training process, our question is: \emph{Can we infer the presence of Differential Privacy in Deep Learning models’ weights?} So, we hypothesize that regardless of the training dataset, architecture and training hyperparameters, the set of DL models trained with and without DP-SGD are separable attending to statistical properties of their model weights. Thus, a meta-classifier should be able to discriminate whether a model employs DP. In simpler terms, our main hypothesis is that differentially private training of a DL model is a property present in its weights and neither it is related to the training dataset nor it is to the architecture of the DL model.

To evaluate our hypothesis, we create an experimental methodology based on a conceptual framework, which formalizes our approach. Our experimental methodology is based on two pillars:

\begin{itemize}
    \item Two sets of 80,000 trained DL models, FCN-Zoo and CNN-Zoo, aimed at providing a train and test grounds to distinguish whether a DL model uses DP, with fully connected and convolutional architectures, respectively. Each comprises 4 subsets of 20,000 trained DL models' weights on the same dataset, half of each subset trained with DP. Both sets of DL models are trained on four relevant image classification datasets, namely MNIST \citep{mnist}, Fashion MNIST \citep{xiao2017fashion}, SVHN \citep{svhn} and CIFAR 10 \citep{svhn}.
    \item Meta-classifiers to discriminate between DL models' weights trained with and without DP for each subset of 20,000 trained DL model weights. Fixed an architecture and without any additional fine-tuning, we use each meta-classifier to predict the presence of DP training in the other 3 subsets of 20,000 trained DL models' weights, to show that the training dataset of the DL model was trained on is not relevant. Furthermore, we remove the assumption of fixing the architecture, to show that the neither the architecture nor the training datasets are relevant when inferring whether a model uses DP in its training process.
\end{itemize}

In addition to this experimental methodology, we formalize the idea of separating DL models according to their usage of DP, that is, we enunciate our theoretical conceptual framework and enunciate our hypotheses of study.

\danii{Employing our experimental methodology, we show that DP imprints DL model weights, so that models trained with and without DP are distinguishable attending to their weights, regardless of the dataset used to train the DL model and its architecture. An ideal property to use in contexts where data privacy is a strict requirement and DP enforcement is required, as it allows any third party to certificate that a model is differentially private. Stated differently, it permits auditing the presence of DP in the training process of a DL model, once it is released. Hopefully, our research will broaden and boost the knowledge about the impact of DP in DL models.}


This paper is structured as follows. First, in Section \ref{sec:related.works}, we introduce all the background knowledge required to understand our conceptual framework and experimental methodology. \daniiiii{Next, Section \ref{sec:formal} introduce the theoretical conceptual framework.} Then, in Section \ref{sec:exp} we enunciate our experimental methodology, consisting of the creation of the datasets, FCN-Zoo and CNN-Zoo, and the training of the meta-classifiers, which extensively test our hypotheses of study. Lastly, in Section \ref{sec:conclu} summarize our findings and point in which directions our work can be further extended to benefit the understanding of DP in the weights of DL models.


\section{\danii{Differential Privacy in Deep Learning and Deep Learning model's weights properties}} \label{sec:related.works}

In this section, we provide the key aspects of the literature required to fully understand the rest of the paper. We introduce DP concepts and provide insight on how previous publications addressed the problem of studying properties of DL models' weights.

\subsection{Differential Privacy with Deep Learning}

The combination of DL and DP offers a comprehensive approach to secure and private DL.



\paragraph{Differential Privacy} It addresses the problem of accessing sensitive data while measuring the consequent exposure or private leakage, stated differently, it manages the fact that accuracy comes at the cost of privacy. 
 
An algorithm $\mathcal{A}$ preserves $\varepsilon$-DP for $\varepsilon > 0$ if for all datasets differing in exactly one element $x,\,y $ and all subset of outputs $\mathcal{O}$ of $\mathcal{A}$ it holds that:

\begin{equation}
	P[\mathcal{A}(x) \in \mathcal{O}] \leq e^{\varepsilon}  P [\mathcal{A}(y) \in \mathcal{O}] 
\end{equation}

If, on the other hand, for $0<\delta< 1$ it holds that:

\begin{equation}
P[\mathcal{A}(x) \in \mathcal{O}] \leq e^{\varepsilon}  P [\mathcal{A}(y) \in \mathcal{O}] + \delta
\end{equation}

then the algorithm possesses the \textit{weaker} property of $(\varepsilon, \delta)$-DP, also known as, approximate DP \citep{dwork2014}. 

DP specifies a ``privacy budget'' given by $\varepsilon$ and $\delta$, where $\varepsilon$ limits the quantity of privacy loss permitted and $\delta$ is the probability of exceeding the privacy budget given by $\varepsilon$.

\paragraph{Deep Learning with Differential Privacy: Differentially Private Stochastic Gradient Descent} By incorporating DP into the Stochastic Gradient Descent algorithm, the Differentially Private Stochastic Gradient Descent (DP-SGD) \citep{Abadi2016} allows the learning of DL models on sensitive datasets while mitigating the risk of exposing individual information. To achieve differential privacy, DP-SGD adds carefully calibrated noise to the calculated gradients. In each iteration of DP-SGD, a batch of training examples is randomly selected from the dataset, just like in the SGD. The contribution to the gradient of each example is limited by clipping the $l_2$ norm of each gradient. The clip value is known as sensitivity. Then noise drawn from a Gaussian distribution is added to the average of the clipped gradients. The amount of noise added to the gradients depends on the sensitivity, the privacy budget ($\varepsilon, \delta$), the size of the batch, the size of the training dataset and the number of training rounds or epochs. In each epoch, the noise is scaled by a factor called the privacy accountant, which keeps track of the remaining privacy budget. \danii{Given that this procedure significantly hurts performance \cite{bagdasaryan2019differential} and since different features have different impacts on the model output, alternative approaches propose to add noise adaptively based on the relevance between different features and the model output \cite{GONG2020131}.}

\dani{\paragraph{Deep Learning with Differential Privacy: accountability} DP is a desirable property of a trained DL model, acquired at training time by applying the DP-SGD. However, once the training is done, to our knowledge, there is no way to check whether the DL model is differentially private. In other words, the model provider has to be trusted. MIA \citep{carlini2021membership} can be used to estimate the DP guarantees of a DL model, however when DP is not present, the privacy guarantees are nonexistent, that is, $\varepsilon=\infty$. However, MIA can only estimate finite DP guarantees, due to the limitations of the Monte Carlo estimation used \citep{measureDP}. Still, \citet{Hyland2019AnES} gave a heuristic argument that SGD itself satisfies DP. Note that, their source of \textit{privacy guarantees} comes from the stochasticity of the SGD and not the random sampling of the batches. Nevertheless, the results of \citet{nasr2021adversary} imply that the gap between measured DP and theoretical DP is almost nonexistent. Their finding might suggest that the privacy guarantees of the SGD itself are vacuous as if they were not, they should be part of the gap between measured DP and theoretical DP.}

\subsection{Deep Learning properties present in model's weights}

The main properties of a trained DL model present in its weights that caught the attention of many researchers can be summarized in two main trends: predicting their performance and predicting the generalization gap, that is, the difference between training and test set performance. \dani{The former approach focuses on either predicting the performance of a trained DL model without the need for any test data, or forecasting the performance of a DL model during the initial stages of its training process. The latter focus on the more general task of forecasting the difference between train and test data performance, that is, the generalization gap of a DL model.}

\paragraph{Predicting performance from weights} When it comes, to predicting the performance of a DL model, just by looking at its weights, \citet{unterthiner2020predicting} showed that it is possible to predict the expected accuracy of a convolutional DL model and that those predictors can rank DL models trained on unseen datasets with different architectures. Similar works presented in \citet{heavytailMartin2020, martin_predicting_2021} show that properties derived from weight matrices correlate well with of DL models in vision and language processing.

Shifting to the field of hyperparameter optimization and neural architecture search, results in similar studies to predict performance from weights. \citet{Streeter2019LearningEL, pmlr-v97-streeter19a} propose procedures that select good hyperparameter values. Based on a few training iterations, \citet{Swersky2014FreezeThawBO, domhan2015} predict the performance of a neural network, to apply early stopping to unsuccessful runs. In the field of neural architecture search, \citep{acceBowen2018, tapas2019} employed analogous methods for selecting candidate architectures, typically relying on hyperparameters, architecture details, dataset information, and performance metrics of comparable architectures for prediction.

\paragraph{Predicting the generalization gap from weights} \citet{predJiang2019} train large convolutional architectures on CIFAR datasets, estimating the minimal distances to the class boundary for every data point within each hidden layer. Utilizing this margin distribution, they employ a linear regressor to forecast generalization gaps. \citet{yak2019towards} builds upon this research by training multiple small, fully connected networks on various iterations of a generated spiral dataset.

Combining both trends, \citet{representacionPesos2021} propose to apply self-supervised learning to create novel representations of DL model weights, that retain enough information to successfully predict the performance and hyperparameters of a DL model, as well as, its generalization gap.

Out of the two main research trends, the overall setting and motivations of \citet{EJRUY20} are similar to the ones in \citet{unterthiner2020predicting}, however instead of predicting the accuracy, they focus on predicting the DL model hyperparameters.

\section{\danii{Theoretical aspects of the conceptual framework: discussion and formal setting}} \label{sec:formal}

\dani{In the following, we establish the main dependent variables used to compute the privacy budget, to explore whether \textit{apparently} independent variables play a role in determining the presence or absence of DP guarantees in DL models. Then, we present a formal introduction to the conceptual framework and present the principal theoretical aspects we will study \danii{with our experimental methodology} in the next section.}

On the one hand, for the DP-SGD, we have a privacy accountant $P$, which considers the batch size $B$, the size of the dataset $S$, the number of training epochs $E$, the probability of exceeding the privacy budget $\delta$ and the noise multiplier $\sigma$, and returns $\varepsilon$ the privacy budget spent in the training process using the DP-SGD. Thus, the trained DL model has $(\varepsilon, \delta)$-DP after E training epochs.

More formally, the privacy accountant is a function\footnote{$\mathds{N}$ and $\mathds{R}_{+}$ stand for the set of Natural numbers and the set of Real positive numbers, respectively.} $P: \mathds{N}^3\times\mathds{R}_{+}^2\rightarrow \mathds{R}_{+}$, given by $P(S, B, E, \sigma, \delta)=\varepsilon$, which computes the approximate minimum $\varepsilon$ for $\delta$ under the assumption of using the Gaussian mechanism with noise scaled to $\sigma$, composed $E$ times, where the probability of an element occurring in a batch is $B/S$, sampled from a Poisson distribution from a dataset of size $S$. Moreover, the composition of the Gaussian mechanism is computed under the composition theorems of Rényi Differential Privacy \citep{renyi} and then converted back to approximate DP \citep{Abadi2016}. It is relevant to note that Poisson sampling is not usually done in training pipelines, but assuming that the data was randomly shuffled, it is believed the actual $\varepsilon$ should be closer to this value than the conservative assumption of an arbitrary data order \citep{ponomareva2023dpfy}.

On the other hand, the standard SGD which takes as arguments the batch size, the size of the dataset and the number of epochs, provides no privacy guarantees, that is, $\varepsilon=\infty$. 

In both situations, with and without DP, when computing the privacy guarantees, it is important to remark that there is a weak dependency with the dataset, its size, but there is no dependency with the dimension of the training data, the architecture of the DL model, neither with the hyperparams of the weights, not even with the train and test performance metrics. All these elements are not considered when computing the approximate DP guarantees. These elements are solely considered when weighing the trade-off between utility and privacy. Then, we wonder if any combination of these privacy independent parameters allow us to infer the presence of DP in DL models, once they are trained, given that to the best of our knowledge, there is no way of achieving it in the literature.

Formally, we investigate if there exists a classifier $F$, which separates the space of trained DL models into models trained with DP and models trained without DP, taking as inputs the models' weights $W$, hyperparameters $\lambda$ and values of the performance metrics $\#P(W)$. Note that, weights themselves depend on the neural architecture $A$, training data $D_{tr}$ and hyperparameters $\lambda$, so we abuse notation and write $W=W(A, D_{tr}, \lambda)$. 

If we assume that $F$ exists, we do not know its actual domain, so initially we consider it is the space composed of the set of weights of trained DL models $\mathcal{W}$, the set of all hyperparameters $\Lambda$ and the set of all the performance metrics values $\#\mathcal{P}(\mathcal{W})$, that is, $F:\mathcal{W}\times\Lambda\times\#\mathcal{P}(\mathcal{W})\rightarrow \{DP, \neg DP\}$. We highlight, two aspects, the former is that the noise multiplier is not included in the set of all hyperparameters as its presence is enough to perform the classification task. The latter is that the considered domain of $F$ is too diverse to tackle experimentally, so we restrict it to a subset $\mathcal{W}'=\mathcal{W}(A', D_{tr}', \cdot) \subset\mathcal{W}$ with the set of weights obtained from a fixed dataset $D_{tr}'$ and a fixed architecture $A'$, that is, $f=F|_{\mathcal{W}'\times\Lambda\times\#\mathcal{P(\mathcal{W}')}}$. Then, we would like to know if our approximated $f$ under that restricted domain is a good approximation for $f_{1}=F|_{\mathcal{W}_1\times\Lambda\times\#\mathcal{P}(\mathcal{W}_{1})}$, where $\mathcal{W}_1=\mathcal{W}_{1}(A', \cdot, \cdot) \subset\mathcal{W}$, or for $f_{2}=F|_{\mathcal{W}_{2}\times\Lambda\times\#\mathcal{P}(\mathcal{W}_{2})}$, where $\mathcal{W}_{2}=\mathcal{W}(\cdot, D_{tr}', \cdot) \subset\mathcal{W}$, that is, $f$ is still a good approximation when we remove either one of the assumptions of fixed architecture and training dataset.

We split our main hypothesis of discriminating between models trained with and without DP into two generalization hypotheses, given a meta-classifier $f:{\mathcal{W}'\times\Lambda\times\#\mathcal{P(\mathcal{W}')}}\rightarrow \{DP, \neg DP\}$ with $\mathcal{W}'=\mathcal{W}(A', D_{tr}', \cdot)$ trained on features from weights with a fixed combination of architecture and dataset, $(A', D_{tr}')$:
\begin{itemize}
    \item \textit{Hypothesis I}: $f$ generalizes well to unseen features from models with architecture $A'$, but training dataset $D_{tr}''$, where $D_{tr}''\neq D_{tr}'$. Stated differently, $f$ is a good approximation of $f^1:\mathcal{W}_1\times\Lambda\times\#\mathcal{P}(\mathcal{W}_{1})\rightarrow \{DP, \neg DP\}$, where $\mathcal{W}_1=\mathcal{W}_{1}(A', D_{tr}'', \cdot)$ for any $D_{tr}''\neq D_{tr}'$.
    \item \textit{Hypothesis II}: $f$ generalizes well to unseen features from models with different architecture $A''$ with $A''\neq A'$ and same training dataset $D_{tr}'$. Stated differently, $f$ is a good approximation of $f_2:\mathcal{W}_2\times\Lambda\times\#\mathcal{P}(\mathcal{W}_{2})\rightarrow \{DP, \neg DP\}$, where $\mathcal{W}_2=\mathcal{W}_{2}(A'', D_{tr}', \cdot)$ for any $A''\neq A'$.
\end{itemize}

Assuming that both, Hypothesis I and II are true, we can combine them and further study the generalization regardless of the dataset and the architecture.

\section{\danii{Experimental methodology for discussing whether DP leaves an imprint in DL model weights}} \label{sec:exp}

In this section, \danii{we describe our experimental methodology} by first building our experimental setup, that is, we build a dataset of trained DL models with multiple datasets and architectures, with and without DP. Then, we proceed to train a meta-classifier $f$ for each fixed combination of architecture and dataset \danii{in order to perform an experimental analysis of our hypotheses\footnote{The code for reproducing the experiments of this section is publicly available at: \url{https://github.com/xehartnort/dp-from-weights}}. Thus, the objective of this section is threefold:} 
\begin{enumerate}
    \item \danii{To build an experimental scenario well-suited to verify our hypotheses. We describe how the datasets of DL models, FCN-Zoo and CNN-Zoo, are created in Section \ref{subsec:the_dataset}}.
    \item To find the most appropriate domain of function $f$ for each combination of architecture and dataset, that is, we explore which features of the FCN-Zoo and CNN-Zoo result in better meta-classifiers in Section \ref{subsec:the_meta_classifier}.
    \item To test whether, without additional fine-tuning, our meta-classifiers $f$ are good approximations of more general meta-classifiers. Particularly, we test our hypothesis on the FCN-Zoo and the CNN-Zoo in Section \ref{subsec:the_transference}.
\end{enumerate}


\subsection{Building the datasets of \danii{Deep Learning models with and without Differential Privacy}: FCN-Zoo and CNN-Zoo} \label{subsec:the_dataset}

Motivated by the necessity of classifying DL models trained with and without DP, we created two datasets of 80,000 trained DL models, the FCN-Zoo and the CNN-Zoo, with the following properties:
\begin{enumerate*}[label=(\roman*)]
    \item The architectures considered for FCN-Zoo and CNN-Zoo are a fully connected network and a convolutional network, respectively. The fully connected network is made of a single hidden layer with 128 neurons. The convolutional network is composed of 4 layers, 3 convolutional layers with kernel of size 3, and stride 2 followed by a global average pooling layers and a fully connected output layer;
    \item 20,000 models are trained on each one of the following popular image classification datasets, MNIST \citep{mnist}, Fashion MNIST \citep{xiao2017fashion}, grayscale SVHN \citep{svhn} and grayscale CIFAR 10 \citep{cifar} and each type of architecture. Half of them, 10,000, are trained with DP and the remaining half without it;
    \item According to the discussion of \citet{unterthiner2020predicting}, for each dataset, we sample 10,000 different hyperparameter configurations chosen independently at random from pre-specified ranges, that is, ensure that each hyperparameter combination is unique, and we do not repeat the same combination with different randoms seeds. We stress that 10,000 hyperparameter configurations are sampled for DP training and another 10,000 are sampled for non-DP training. \dani{Regardless of the architecture of the model and the training dataset, each configuration is sampled from the hyperparameter range described in Table \ref{tab:hyp_hyp}};
    \item For each trained model, we store its raw weights, accuracy and loss at train and test sets in the last epoch and hyperparameters used. Additionally, if DP training is considered, we store the $(\varepsilon, \delta)$ value.
\end{enumerate*}

\begin{adjustwidth}{-2.5 cm}{-2.5 cm}\centering\begin{threeparttable}[!htb]
\scriptsize
\begin{tabular}{lll}\toprule
\textbf{Hyperparameter name} &\textbf{Range of values} \\\midrule
\textit{Training size fraction} &values from 0.3 to 1 with step size 0.05 \\\midrule
\textit{Batch size} &values from $2^5$ to $2^{11}$ with step size 1 \\\midrule
\textit{Number of epochs} &5 if the architecture is fully connected, 18 otherwise \\\midrule
\textit{L2 clip of gradient norms} &100 values equally spaced from 0.1 to 1.5. \\\midrule
\textit{Noise multiplier} & \makecell[l]{if DP training is present 10,000 values equally\\ spaced from $10^{-3}$ to 1.5. Otherwise, it is 0} \\\midrule
\textit{Optimizer} &\makecell[l]{Stochastic Gradient Descent (SGD) or Adam \cite{adam}.\\ if DP training is present, we use their DP counterparts} \\\midrule
\textit{Learning rate} &10,000 values equally spaced from $10^{-3}$ to 0.1 \\\midrule
\textit{Activation function} &\makecell[l]{Hyperbolic Tangent (Tanh) or Rectified Linear \\ unit (ReLu).} \\\midrule
\makecell[l]{\textit{Weight initialization scheme}} & \makecell[l]{Glorot normal initializer \citep{glorot}, normal distribution, \\truncated normal distribution, an orthogonal matrix \citep{saxe2014a}, \\ He normal initialization \citep{Henormal} }\\\midrule
\makecell[l]{\textit{Weight initialization standard deviation}} &10,000 values equally spaced from 0.1 to 0.5 \\
\bottomrule
\end{tabular}
\caption{Hyperparameters and their ranges considered for training 10,000 DL models, regardless of the architecture and the training dataset.}\label{tab:hyp_hyp}
\end{threeparttable}\end{adjustwidth}

\begin{figure}[h!]
\centering
\begin{subfigure}{.5\textwidth}
  \centering
    \includegraphics[width=\linewidth]{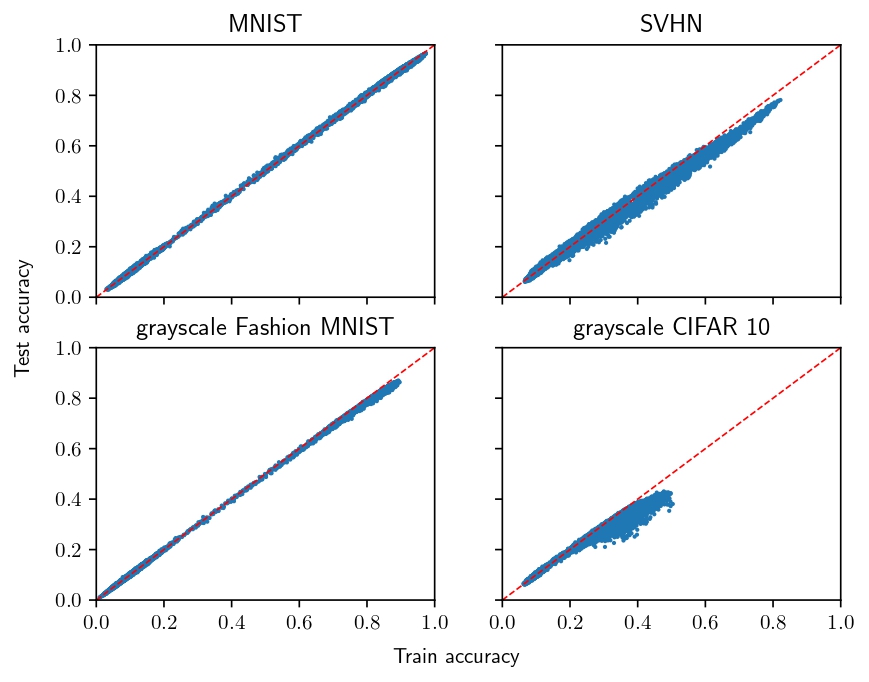}
    \subcaption{Train and test accuracies in the FCN-Zoo}
\end{subfigure}%
\begin{subfigure}{.5\textwidth}
  \centering
    \includegraphics[width=\linewidth]{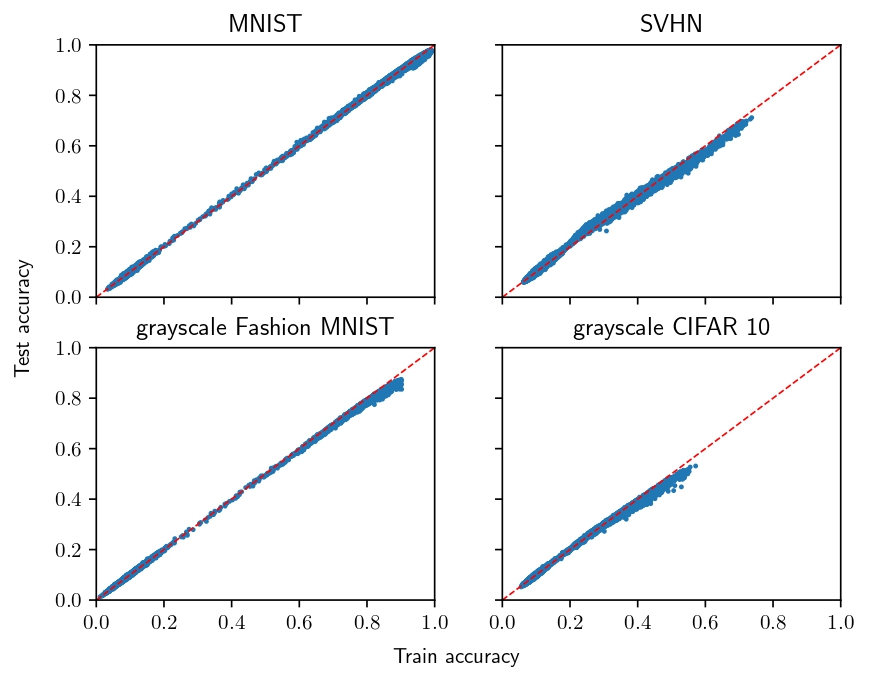}
    \subcaption{Train and test accuracies in the CNN-Zoo}
\end{subfigure}
\caption{From left to right, each point represents the relation of train/test accuracies of an individual model trained without DP, for each of the four image classification datasets in the FCN-Zoo and CNN-Zoo, respectively. The red dashed line represents the ideal relationship between train and test accuracies, that is, each point under the line presents some degree of overfitting while each point over the line presents some degree of generalization.}
\label{fig:acc}
\end{figure}

In Figure \ref{fig:acc}, we observe that the train-test accuracies of the models in CNN-Zoo and FCN-Zoo are far from state-of-the-art for CIFAR 10 and SVHN, nevertheless, it has above 90\% accuracies in MNIST and Fashion MNIST. The red dashed line represents the ideal relationship between train and test accuracies, that is, each point under the line presents some degree of overfitting while each point over the line presents some degree of generalization. Note that, the smallest convolutional model, achieving above 90\% test accuracy on CIFAR 10, requires multiple orders more of parameters \citep{DB15a}. Furthermore, we are considering the grayscale version of CIFAR 10 and SVHN, which also hinders test accuracy, but allows us to re-use the same architecture for all 4 datasets. The distribution of the train and test accuracy of the model without DP do not show any sign of overfitting in the CNN Zoo, but the same does not hold true for SVHN and CIFAR 10 in the FCN Zoo, where there are small signs of overfitting. 

\begin{figure}[h!]
\centering
\begin{subfigure}{.5\textwidth}
  \centering
    \includegraphics[width=\linewidth]{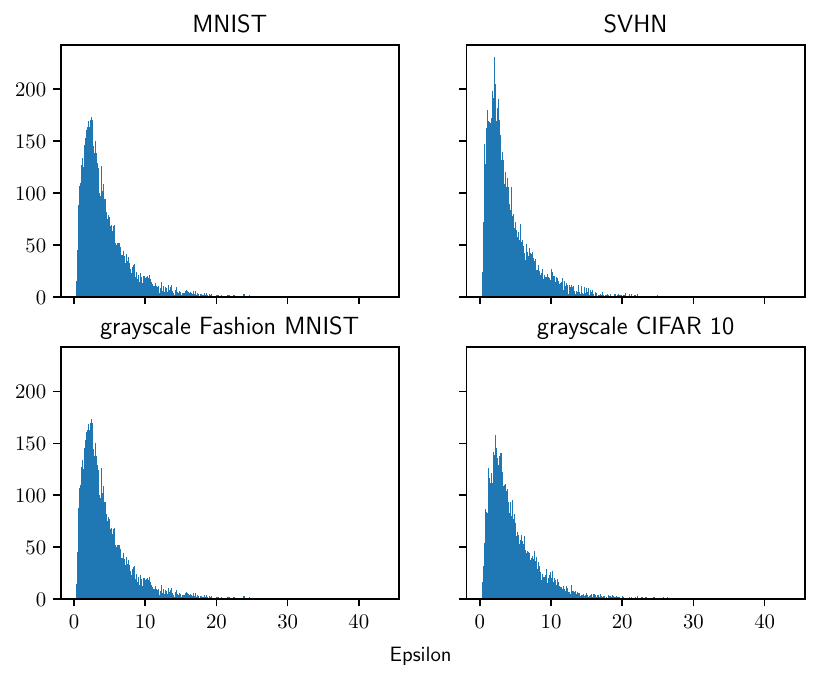}
    \subcaption{Histogram of $\varepsilon$ values in the FCN-Zoo}
\end{subfigure}%
\begin{subfigure}{.5\textwidth}
  \centering
    \includegraphics[width=\linewidth]{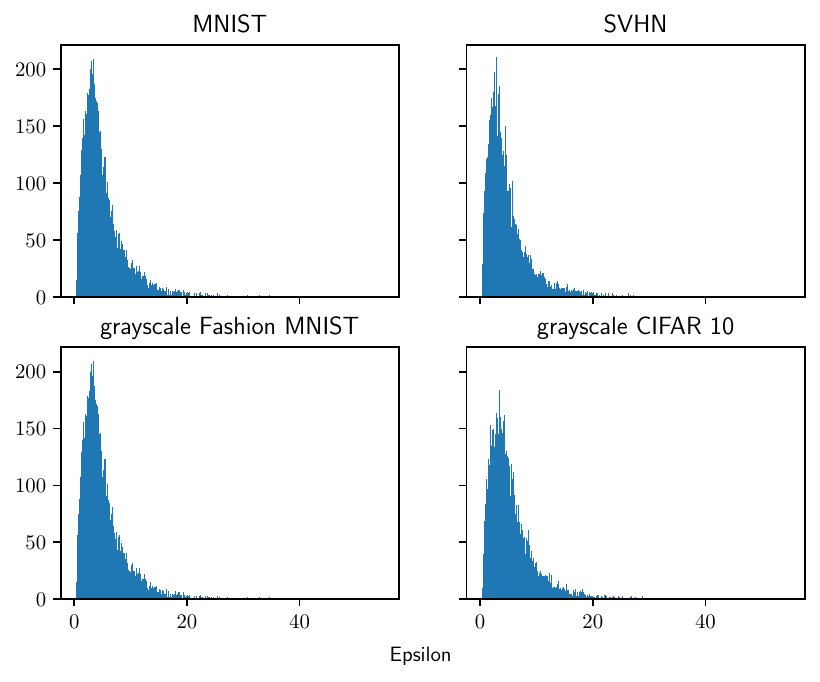}
    \subcaption{Histogram of $\varepsilon$ values in the CNN-Zoo}
\end{subfigure}
\caption{The distribution of $\varepsilon$ values across four image classification datasets in the FCN-Zoo and CNN-Zoo, respectively.}
\label{fig:eps_fixed}
\end{figure}

Figure \ref{fig:eps_fixed} shows the distribution of epsilon values, which is similar for CNN-Zoo and FCN-Zoo. The main difference, apart from the architecture, is that in the CNN-Zoo, models are trained for 18 epochs and in the FCN-Zoo they are trained for 5 epochs. Clearly, most epsilon values are conglomerated around the interval $[0,10]$, which comes handy, as in practice when applying DP it is desirable to achieve a low $\varepsilon$.

\subsection{Training and evaluating meta-classifiers} \label{subsec:the_meta_classifier}

We choose as a meta-classifier LightGBM \citep{lightgbm}, a Gradient Boosting Machine decision tree model, minimizing binary cross entropy motivated by its usage in \citet{unterthiner2020predicting}. It has many hyperparameters and early experiments strongly suggested that it is important to tune them. For each subset of 20,000 models trained on a fixed dataset and architecture, half of them with DP, we create an 75\%-25\% train-test split. Then, on the train split, we perform hyperparameter selection by evaluating 500 unique hyperparameter configurations sampled randomly and independently, from pre-specified ranges detailed in \dani{Table \ref{tab:lgbm}}. The best model is selected based on 3-fold cross-validation performed on the training split, then the evaluation of the best model is done using the test split only once.

\begin{table}[!htp]\centering
\scriptsize
\begin{tabular}{lll}\toprule
LightGBM hyperparameter name &Range of values \\\midrule
\textit{num\_leaves} &sampled uniformly from $[20, 10^{4}]$ \\
\textit{max\_depth} &sampled log-uniformly from $[5, 15]$ \\
\textit{learning\_rate} &sampled log-uniformly from $[10^{-2}, 10^{-1}]$ \\
\textit{max\_bin} &sampled uniformly from the set $\{2^6-1, 2^7-1, 2^8-1\}$ \\
\textit{min\_child\_weight} &sampled uniformly from the set $\{1,2,3,4,5\}$ \\
\textit{reg\_lambda} &sampled uniformly from $[10^{-3}, 100]$ \\
\textit{ref\_alpha} &sampled uniformly from $[10^{-6}, 5]$ \\
\textit{subsample} &sampled uniformly from $\{0.1,0.2,...,1\}$ \\
\textit{subsample\_freq} &fixed to 1 \\
\textit{colsample\_bytree} &sampled log-uniformly from $[10^{-2}, 10^{-1}]$ \\
\bottomrule
\end{tabular}
\caption{LightGBM hyperparameter ranges considered when training meta-classifiers. We refer to the LightGBM documentation for the concrete meaning of the parameters.}\label{tab:lgbm}
\end{table}

It is important to note that we want to study if we can detect the presence of DP training in a model, ideally, regardless of the dataset and the architecture. Consequentially, instead of considering the full raw vector of concatenated weights and bias as input in the classification task, we summarize each layer and bias weights using simple statistics properties such as mean, standard deviation and quantiles 0, 25, 50, 75, 100. These statistical properties allow us to reuse the meta-classifiers with different architectures in the following sections. Additionally, we consider the following information as features to train the meta-classifiers:
\begin{itemize}
    \item \dani{\textit{$\#P$: performance metrics values}. They consist in the accuracy and cross-entropy values from the DL models obtained in train and test splits of the corresponding image classification dataset.}
    \item \dani{\textit{$\lambda$: DL models hyperparameters}. Hyperparameters taken into consideration are: sampling ratio, number of optimization steps, learning rate, activation function of the hidden layers, weight initialization scheme and optimizer. Table \ref{tab:hyp_hyp} details the range of values considered. Aside from those, we include the following hyperparameters used in the $(\varepsilon, \delta)$ computation:}
\begin{itemize}
    \item \dani{\textit{Sampling ratio}, which is the ratio: batch size / number of training samples.}
    \item \dani{\textit{Number of optimization steps}, the number of global optimization steps taken.}
\end{itemize}

\end{itemize}

\dani{When DP training is enabled, we found the $\varepsilon$ distribution to have a long tail which included values orders of magnitude higher than 10, which are considered to provide low or negligible privacy guarantees. To remove such a long tail, we limited the epsilon values to 10 and over sampled the batch size, noise multiplier range and train fraction to 25,000 elements and then removed randomly 15,000 configurations with epsilon in range $[1, 5]$. Lastly, we reshuffled our hyperparameters, achieving a lighter tail while preserving $\varepsilon$ values around 10, as shown in Figure \ref{fig:eps_fixed}.}

\begin{table}[!htp]\centering
\scriptsize
\begin{tabular}{lccccc}\toprule
\makecell{FCN-Zoo\\features} &MNIST &\makecell{Fashion\\MNIST} &\makecell{Grayscale\\SVHN} &\makecell{Grayscale\\CIFAR 10} \\\midrule
$\lambda$ &0.500 &0.500 &0.500 &0.500 \\
$\#P$ &0.701 &0.722 &0.761 &0.721 \\
$\lambda+\#P$ &0.839 &0.878 &0.864 &0.856 \\
$W_1$ &0.968 &0.957 &0.972 &0.966 \\
$W_2$ &0.992 &0.986 &0.982 &0.992 \\
$W_2+\#P$ &0.999 &0.999 &0.995 &0.999 \\
$W_2+\lambda$ &0.996 &0.993 &0.995 &0.997 \\
$W_2+\lambda+\#P$ &0.999 &0.999 &0.998 &0.999 \\
$W_1+W_2$ &0.998 &0.998 &0.997 &0.997 \\
\end{tabular}
\begin{tabular}{lccccc}\toprule
\makecell{CNN-Zoo\\features} &MNIST &\makecell{Fashion \\ MNIST} & \makecell{Grayscale \\ SVHN} &\makecell{Grayscale \\ CIFAR 10} \\\midrule
$\lambda$ &0.500 &0.500 &0.500 &0.500 \\
$\#P$ &0.836 &0.854 &0.882 &0.835 \\
$\lambda +\#P$ &0.920 &0.932 &0.944 &0.905 \\
$W_1$ &0.926 &0.904 &0.940 &0.918 \\
$W_2$ &0.923 &0.930 &0.909 &0.912 \\
$W_3$ &0.941 &0.939 &0.923 &0.926 \\
$W_4$ &0.971 &0.975 &0.968 &0.976 \\
$W_4 + \#P$ &0.999 &0.999 &0.995 &0.998 \\
$W_4 + \lambda$ &0.993 &0.992 &0.993 &0.996 \\
$W_4 + \lambda + \#P$ &0.999 &0.999 &0.998 &0.999 \\
$\sum_{i=1}^{4} W_i$ &0.997 &0.995 &0.996 &0.998 \\
\bottomrule
\end{tabular}
\caption{Accuracy of meta-classifiers trained on multiple combinations of the features present in FCN-Zoo and CNN-Zoo, where $\lambda$ stands for hyperparameters,$\#P$ for values of performance metrics, $W_i$ for weight stats of layer $i$ and $+$ for the union of sets.}\label{tab:meta}
\end{table}

Table \ref{tab:meta} incorporates the accuracy score of the meta-classifiers when trained with multiple combinations of features, namely, weights statistics of model layers $W_i$, hyperparameters $\lambda$ , and performance metrics values $\#P$, in FCN-Zoo and CNN-Zoo, respectively. We highlight that hyperparameters, $\lambda$, alone are not enough to infer which models are trained with DP. It also suggests that the selection of the hyperparameters does not introduce any bias that significantly eases the meta-classification tasks. We find that $\#P$ achieves a great accuracy score, hinting the fact that DP significantly hinders performance \citep{bagdasaryan2019differential}. 

In both, FCN-Zoo and CNN-Zoo every layer individually $W_i$ is enough to achieve a high classification accuracy, specially the last one. Indeed, their combination achieves one of the highest accuracies. We also highlight that the union of the last layer of weight statistics and the performance metric values, provides a slight boost of accuracy when compared to the statistics of the last layer alone. We also find interesting that nor the low accuracy values neither the small signs of overfitting observed in Figure \ref{fig:acc} seem to increase the difficulty of the meta-classification task.

\subsection{Generalization properties of meta-classifiers} \label{subsec:the_transference}

In this section, we test the generalization capabilities of the meta-classifiers trained in Section \ref{subsec:the_meta_classifier}. We begin testing the \textit{Hypothesis I}, using the insights obtained previously, that is, the weight statistics of the last layer, the union of all of them and the performance values are the best features to train the meta-classifiers. Then, we continue testing the \textit{Hypothesis II} and its combination with \textit{Hypothesis I}, using the same insights. As a result, we obtain which features allow the meta-classifier to generalize better and the complete generalization capabilities of the meta-classifiers.

\begin{adjustwidth}{-2.5 cm}{-2.5 cm}\centering\begin{threeparttable}[!htb]
\scriptsize
\begin{tabular}{crrrr|rrrrr}\toprule
\makecell{FCN-Zoo\\features} &\multicolumn{4}{c}{$\#P$} &\multicolumn{4}{c}{$\#P+\lambda$} \\\cmidrule{1-9}
&MNIST &\makecell{Fashion\\MNIST} &\makecell{Grayscale\\SVHN} &\makecell{Grayscale\\CIFAR 10} &MNIST &\makecell{Fashion\\MNIST} &\makecell{Grayscale\\SVHN} &\makecell{Grayscale\\CIFAR 10} \\\midrule
MNIST &- &0.719 &0.531 &0.524 &- &0.836 &0.502 &0.616 \\
\makecell{Fashion\\MNIST} &0.670 &- &0.529 &0.500 &0.835 &- &0.437 &0.427 \\
\makecell{Grayscale\\SVHN} &0.500 &0.487 &- &0.735 &0.584 &0.551 &- &0.830 \\
\makecell{Grayscale\\CIFAR 10} &0.480 &0.467 &0.686 &- &0.599 &0.555 &0.757 &- \\
\end{tabular}
\begin{tabular}{crrrr|rrrrr}\toprule
\makecell{FCN-Zoo\\features} &\multicolumn{4}{c}{$W_1$} &\multicolumn{4}{c}{$W_2$} \\\cmidrule{1-9}
&MNIST &\makecell{Fashion\\MNIST} &\makecell{Grayscale\\SVHN} &\makecell{Grayscale\\CIFAR 10} &MNIST &\makecell{Fashion\\MNIST} &\makecell{Grayscale\\SVHN} &\makecell{Grayscale\\CIFAR 10} \\\midrule
MNIST &- &0.908 &0.857 &0.849 &- &0.952 &0.928 &0.912 \\
\makecell{Fashion\\MNIST} &0.943 &- &0.893 &0.883 &0.973 &- &0.910 &0.897 \\
\makecell{Grayscale\\SVHN} &0.696 &0.708 &- &0.937 &0.908 &0.888 &- &0.932 \\
\makecell{Grayscale\\CIFAR 10} &0.631 &0.633 &0.950 &- &0.888 &0.873 &0.937 &- \\
\bottomrule
\end{tabular}
\caption{Accuracy of meta-classifiers trained on specific FCN-Zoo features from models trained on a fixed dataset from the first column, applied to FCN-Zoo features from unseen models trained on the datasets listed in rows. The features considered for each case are detailed in the top row of each sub-table.}\label{tab:hypi_fcn_part1}
\end{threeparttable}\end{adjustwidth}

\paragraph{Hypothesis I: meta-classifiers generalizations regardless of the image classification dataset} Tables \ref{tab:hypi_fcn_part1} and \ref{tab:hypi_cnn_part1} explore the accuracy of meta-classifiers trained on simple features from the FCN-Zoo and CNN-Zoo, respectively, applied to unseen features. Note that each feature is extracted from models with the same architecture but trained on different datasets. Compared to Table \ref{tab:meta}, we find that the accuracy gap of using $\# P$ or $\# P+\lambda$ and $W_3$ or $W_4$ as training features is wider, showing that the weight statistics of the last layers provide greater generalization properties to the meta-classifiers.

\begin{adjustwidth}{-2.5 cm}{-2.5 cm}\centering\begin{threeparttable}[!htb]
\scriptsize
\begin{tabular}{ccccc|ccccc}\toprule
\makecell{CNN-Zoo\\features}&\multicolumn{4}{c}{$\#P$} &\multicolumn{4}{c}{$\lambda +\#P$} \\\cmidrule{1-9}
&MNIST&\makecell{Fashion\\MNIST}&\makecell{Grayscale\\SVHN}&\makecell{Grayscale\\CIFAR 10}&MNIST &\makecell{Fashion\\MNIST}&\makecell{Grayscale\\SVHN}&\makecell{Grayscale\\CIFAR 10}\\\midrule
MNIST &- &0.829 &0.633 &0.598 &- &0.893 &0.657 &0.628 \\
\makecell{Fashion\\MNIST} &0.829 &- &0.642 &0.615 &0.913 &- &0.649 &0.621 \\
\makecell{Grayscale\\SVHN} &0.686 &0.629 &- &0.822 &0.692 &0.630 &- &0.835 \\
\makecell{Grayscale\\CIFAR 10} &0.641 &0.579 &0.820 &- &0.688 &0.617 &0.864 &- \\
\end{tabular}
\begin{tabular}{ccccc|ccccc}\toprule
\makecell{CNN-Zoo\\features}&\multicolumn{4}{c}{$W_3$} &\multicolumn{4}{c}{$W_4$} \\\cmidrule{1-9}
&MNIST&\makecell{Fashion\\MNIST}&\makecell{Grayscale\\SVHN}&\makecell{Grayscale\\CIFAR 10}&MNIST &\makecell{Fashion\\MNIST}&\makecell{Grayscale\\SVHN}&\makecell{Grayscale\\CIFAR 10}\\\midrule
MNIST &- &0.897 &0.877 &0.883 &- &0.908 &0.887 &0.908 \\
\makecell{Fashion\\MNIST} &0.920 &- &0.889 &0.904 &0.898 &- &0.897 &0.891 \\
\makecell{Grayscale\\SVHN} &0.885 &0.876 &- &0.874 &0.902 &0.925 &- &0.897 \\
\makecell{Grayscale\\CIFAR 10} &0.908 &0.906 &0.891 &- &0.867 &0.867 &0.874 &- \\
\bottomrule
\end{tabular}
\caption{Accuracy of meta-classifiers trained on specific CNN-Zoo features from models trained on a fixed dataset from the first column, applied to CNN-Zoo features from unseen models trained on the datasets listed in rows. The features considered for each case are detailed in the top row of each sub-table.}\label{tab:hypi_cnn_part1}
\end{threeparttable}\end{adjustwidth}

\dani{Additionally, we are interested in showing whether the inclusion of $\#P$, $\lambda$ and the union of all weight statistics significantly improves the generalization properties of the meta-classifiers, as it does improve accuracy in Table \ref{tab:meta}. The accuracy values reported when using more complex features are presented in Tables \ref{tab:extra_1} and \ref{tab:extra_2}, for FCN-Zoo and CNN-Zoo, respectively. The inclusion of the hyperparameters values $\lambda$ to the weight statistics of the last layer $W_{-1}$\footnote{Where $W_{-1}=W_2$ for the FCN-Zoo and $W_{-1}=W_4$ for the CNN-Zoo.}, achieves the highest results overall in each table. While the union of all weight statistics $\sum W_i$ achieves slightly smaller scores but still, they present an improvement over all the accuracy scores shown in Tables \ref{tab:hypi_fcn_part1} and \ref{tab:hypi_cnn_part1}. Thus, we can confirm that there is a boost in accuracy when more complex features are used to train the meta-classifiers.}

\dani{We can conclude that for both, the FCN-Zoo and the CNN-Zoo, the features that achieve the best accuracy scores overall are the union of the hyperparameters and the weight statistics of the last layer $\lambda + W_{-1}$ followed by the union of weight statistics $\sum W_i$.}

To summarize, our results allow us to prove that Hypothesis I is correct and the smallest set of features that verifies it with the highest accuracy are the weight statistics from the last layers. Overall, we find that meta-classifiers trained on features from any dataset of the tuples (Fashion MNIST, MNIST) and (SVHN, CIFAR 10), achieve the best generalization accuracy among them.

\begin{adjustwidth}{-2.5 cm}{-2.5 cm}\centering\begin{threeparttable}[!htb]
\scriptsize
\begin{tabular}{crrrr|rrrrr}\toprule
\makecell{FCN-Zoo\\features} &\multicolumn{4}{c}{$W_2+\#P$} &\multicolumn{4}{c}{$W_2+\lambda$} \\\cmidrule{1-9}
&MNIST &\makecell{Fashion\\MNIST} &\makecell{Grayscale\\SVHN} &\makecell{Grayscale\\CIFAR 10} &MNIST &\makecell{Fashion\\MNIST} &\makecell{Grayscale\\SVHN} &\makecell{Grayscale\\CIFAR 10} \\\midrule
MNIST &- &0.995 &0.868 &0.849 &- &0.967 &0.926 &0.951 \\
\makecell{Fashion\\MNIST} &0.953 &- &0.877 &0.868 &0.993 &- &0.922 &0.924 \\
\makecell{Grayscale\\SVHN} &0.749 &0.717 &- &0.994 &0.927 &0.899 &- &0.989 \\
\makecell{Grayscale\\CIFAR 10} &0.779 &0.745 &0.974 &- &0.887 &0.879 &0.963 &- \\
\end{tabular}
\begin{tabular}{crrrr|rrrrr}\toprule
\makecell{FCN-Zoo\\features} &\multicolumn{4}{c}{$W_2 +\#P + \lambda$} &\multicolumn{4}{c}{$W_1+W_2$} \\\cmidrule{1-9}
&MNIST &\makecell{Fashion\\MNIST} &\makecell{Grayscale\\SVHN} &\makecell{Grayscale\\CIFAR 10} &MNIST &\makecell{Fashion\\MNIST} &\makecell{Grayscale\\SVHN} &\makecell{Grayscale\\CIFAR 10} \\\midrule
MNIST &- &0.996 &0.920 &0.919 &- &0.984 &0.926 &0.917 \\
\makecell{Fashion\\MNIST} &0.955 &- &0.875 &0.863 &0.996 &- &0.969 &0.974 \\
\makecell{Grayscale\\SVHN} &0.747 &0.716 &- &0.993 &0.879 &0.895 &- &0.985 \\
\makecell{Grayscale\\CIFAR 10} &0.797 &0.769 &0.976 &- &0.967 &0.957 &0.982 &- \\
\bottomrule
\end{tabular}
\caption{Accuracy of applying meta-classifiers trained on specific FCN-Zoo features from models trained on a fixed dataset from the first column, to FCN-Zoo features from unseen models trained on the datasets listed in rows. The FCN-Zoo features considered when training meta-classifiers are: \textit{weight statistics of the last layer and performance values} $W_2+\#P$, \textit{weight statistics of the last layer and hyperparameters} $W_2+\lambda$, \textit{weight statistics of the last layer, hyperparameters and performance values}  $W_2+\#P+\lambda$, and \textit{the union of weight statistics from all layers} $W_1 + W_2$.}\label{tab:extra_1}
\end{threeparttable}\end{adjustwidth}

\begin{adjustwidth}{-2.5 cm}{-2.5 cm}\centering\begin{threeparttable}[!htb]
\scriptsize
\begin{tabular}{crrrr|rrrrr}\toprule
\makecell{CNN-Zoo\\features}&\multicolumn{4}{c}{$W_4 + \#P$} &\multicolumn{4}{c}{$W_4+\lambda$} \\\cmidrule{1-9}
&MNIST &\makecell{Fashion\\MNIST} &\makecell{Grayscale\\SVHN} &\makecell{Grayscale\\CIFAR 10} &MNIST &\makecell{Fashion\\MNIST} &\makecell{Grayscale\\SVHN} &\makecell{Grayscale\\CIFAR 10} \\\midrule
MNIST &- &0.992 &0.875 &0.872 &- &0.964 &0.951 &0.972 \\
\makecell{Fashion\\ MNIST} &0.997 &- &0.866 &0.858 &0.953 &- &0.953 &0.956 \\
\makecell{Grayscale\\ SVHN} &0.861 &0.790 &- &0.993 &0.958 &0.957 &- &0.976 \\
\makecell{Grayscale\\ CIFAR 10} &0.886 &0.809 &0.980 &- &0.931 &0.927 &0.942 &- \\
\end{tabular}
\begin{tabular}{crrrr|rrrrr}\toprule
\makecell{CNN-Zoo\\features}&\multicolumn{4}{c}{$W_4 + \#P +\lambda$} &\multicolumn{4}{c}{$\sum_{i=1}^4 W_i$} \\\cmidrule{1-9}
&MNIST &\makecell{Fashion\\MNIST} &\makecell{Grayscale\\SVHN} &\makecell{Grayscale\\CIFAR 10} &MNIST &\makecell{Fashion\\MNIST} &\makecell{Grayscale\\SVHN} &\makecell{Grayscale\\CIFAR 10} \\\midrule
MNIST &- &0.994 &0.861 &0.845 &- &0.965 &0.954 &0.959 \\
\makecell{Fashion\\ MNIST} &0.998 &- &0.870 &0.861 &0.976 &- &0.966 &0.955 \\
\makecell{Grayscale\\ SVHN} &0.867 &0.793 &- &0.995 &0.947 &0.949 &- &0.969 \\
\makecell{Grayscale\\ CIFAR 10} &0.885 &0.815 &0.981 &- &0.944 &0.935 &0.965 &- \\
\bottomrule
\end{tabular}
\caption{Accuracy of applying meta-classifiers trained on specific CNN-Zoo features from models trained on a fixed dataset from the first column, to CNN-Zoo features from unseen models trained on the datasets listed in rows. The CNN-Zoo features considered when training meta-classifiers are: \textit{weight statistics of the last layer and performance values} $W_4+\#P$,\textit{ weight statistics of the last layer and hyperparameters} $W_4+\lambda$,\textit{ weight statistics of the last layer}, \textit{hyperparameters and performance values} $W_4+\#P+\lambda$, and \textit{the union of weight statistics from all layers} $\sum_{i=1}^4 W_i$.}\label{tab:extra_2}
\end{threeparttable}\end{adjustwidth}

\paragraph{Hypothesis II: meta-classifiers generalization regardless of the architecture and beyond} We need to train the meta-classifiers on features obtained from models with different architectures. Therefore, such features should have similar meaning and input size, which leave us with hardly any options, namely the weight statistics of the last layers and the performance metrics values. We also explore the combination of Hypothesis I and II to fully discover the generalization capabilities of the meta-classifiers. 

Tables \ref{tab:fcn_2_dnn} and \ref{tab:dnn_2_cnn} show the generalization capabilities in terms of classification accuracy of meta-classifiers trained on features from models with trained on a fixed architecture and training dataset, applied to features from models trained with different architecture. Particularly, the accuracy values of the diagonals correspond to the situation detailed in the Hypothesis II. The remaining values, show that the generalization capabilities of the meta-classifiers also extend when both Hypothesis I and II are considered simultaneously, that is, the meta-classifiers generalize well to unseen features from models with different training dataset and architecture. We highlight that  the generalization is remarkably higher when the meta-classifiers are trained on features from convolutional models, that is, Table \ref{tab:dnn_2_cnn} accuracies are higher than Table \ref{tab:fcn_2_dnn}. Surprisingly, in both tables the inclusion of $\#P$ produces mixed results and in most cases it does not increase accuracy in the diagonals, where the Hypothesis II is tested. In Table \ref{tab:fcn_2_dnn}, only the meta-classifier trained with features from models trained in SVHN achieves higher accuracies when $\#P$ is considered. The same holds for CIFAR 10 in Table~\ref{tab:dnn_2_cnn}.

Our results allow us to prove that Hypothesis II is correct as well as the combination of both Hypothesis I and II, that is, the meta-classifiers generalize regardless of the training dataset and architecture of the features.

\begin{adjustwidth}{-2.5 cm}{-2.5 cm}\centering\begin{threeparttable}[!htb]
\scriptsize
\begin{tabular}{ccccc|ccccc}\toprule
 &\multicolumn{4}{c}{$W_{-1}$} &\multicolumn{4}{c}{$W_{-1} + \#P$} \\\cmidrule{1-9}
&MNIST &\makecell{Fashion\\MNIST} &\makecell{Grayscale\\SVHN} &\makecell{Grayscale\\CIFAR 10} &MNIST &\makecell{Fashion\\MNIST} &\makecell{Grayscale\\SVHN} &\makecell{Grayscale\\CIFAR 10} \\\midrule
MNIST &0.724 &0.711 &0.783 &0.708 &0.710 &0.697 &0.769 &0.704 \\
\makecell{Fashion\\MNIST} &0.789 &0.778 &0.842 &0.794 &0.777 &0.770 &0.833 &0.768 \\
\makecell{Grayscale\\SVHN} &0.789 &0.803 &0.829 &0.784 &0.831 &0.837 &0.853 &0.831 \\
\makecell{Grayscale\\CIFAR 10} &0.841 &0.844 &0.854 &0.850 &0.829 &0.834 &0.841 &0.838 \\
\bottomrule
\end{tabular}
\caption{Accuracy of meta-classifiers trained on FCN-Zoo feature $W_2$ applied to CNN-Zoo feature $W_4$. To avoid confusion, $W_2$ and $W_4$ are noted as $W_{-1}$. Where the meta-classifiers are trained on features from fully connected models trained on datasets listed in the first column and applied, without any fine-tuning, to features from convolutional models trained on datasets listed in rows.}\label{tab:fcn_2_dnn}
\end{threeparttable}\end{adjustwidth}

\begin{adjustwidth}{-2.5 cm}{-2.5 cm}\centering\begin{threeparttable}[!htb]
\scriptsize
\begin{tabular}{ccccc|ccccc}\toprule
 &\multicolumn{4}{c}{$W_{-1}$} &\multicolumn{4}{c}{$W_{-1} + \#P$} \\\cmidrule{1-9}
&MNIST &\makecell{Fashion\\MNIST} &\makecell{Grayscale\\SVHN} &\makecell{Grayscale\\CIFAR 10} &MNIST &\makecell{Fashion\\MNIST} &\makecell{Grayscale\\SVHN} &\makecell{Grayscale\\CIFAR 10} \\\midrule
MNIST &0.884 &0.875 &0.857 &0.863 &0.887 &0.878 &0.865 &0.867 \\
\makecell{Fashion\\MNIST} &0.856 &0.861 &0.844 &0.853 &0.854 &0.859 &0.849 &0.855 \\
\makecell{Grayscale\\SVHN} &0.842 &0.843 &0.797 &0.784 &0.830 &0.834 &0.793 &0.765 \\
\makecell{Grayscale\\CIFAR 10} &0.884 &0.871 &0.859 &0.870 &0.895 &0.879 &0.868 &0.872 \\
\bottomrule
\end{tabular}
\caption{Accuracy of meta-classifiers trained on CNN-Zoo feature $W_4$ applied to FCN-Zoo feature $W_2$. To avoid confusion, $W_4$ and $W_2$ are noted as $W_{-1}$. Where the meta-classifiers are trained on features from convolutional models trained on datasets listed in the first column and applied, without any fine-tuning, to features from fully connected models trained on datasets listed in rows.}\label{tab:dnn_2_cnn}
\end{threeparttable}\end{adjustwidth}

\section{Conclusions and Future work} \label{sec:conclu}

This work contributes to deepening the understanding of the impact of DP on DL models, that is, how DP imprints the weights of DL models, regardless of the training dataset and the architecture of the model. More specifically, we find this property useful to certificate the presence of DP training in a DL model. 

Our experimental methodology has led us to answer the question: \emph{Can we infer the presence of Differential Privacy in Deep Learning models’ weights?} Yes, it is possible to acknowledge the presence of DP in the weights of a DL model. Furthermore, this presence is knowledgeable even if vital parts of a DL model such as its architecture and its training dataset vary, showing that it is a general property of DL models. A useful property to provide accountability of DP training of a DL model, when DP is required due to strict data privacy requirements. 

Additionally, we contribute with two datasets, the FCN-Zoo and the CNN-Zoo. To our knowledge, these are the first datasets to include both models trained with and without DP, providing a great starting point to this interesting direction of research.


Hence, our contributions help to broaden the knowledge about the impact of DP in DL models, certificate it and can hopefully boost the research of DP-based solutions, working towards more secure DL.

Future work will focus on testing more extensively the discovered properties, that is, testing whether our hypotheses hold with attention-based architectures in more diverse tasks such as natural language modelling. Furthermore, we will explore the usage of state-of-the-art machine learning interpretability techniques to provide a wider understanding of the decisions made by the meta-classifiers, thus providing a more significant value to certificating the presence of DP training.

\section{Acknowledgments}

This research work is partially supported by the R\&D\&I, Spain grants PID2020-119478GB-I00 and, PID2020-116118GA-I00 funded by MCIN/AEI/\-10.13039/501100011033.

\bibliography{mybibfile}

\bibliographystyle{unsrtnat}

\end{document}